\documentclass{article}
\usepackage{ijcai22}

\usepackage{times}

\usepackage{soul}
\usepackage{url}
\usepackage[hidelinks]{hyperref}
\usepackage[usenames,dvipsnames]{xcolor}
\usepackage[utf8]{inputenc}
\usepackage[small]{caption}
\usepackage{graphicx}
\usepackage{xspace}
\usepackage{bold-extra}
\usepackage{paralist, tabularx}
\usepackage{subcaption}
\usepackage{multicol, multirow}
\usepackage{amsmath}
\usepackage{dsfont}
\usepackage{booktabs}
\hypersetup{colorlinks=true,allcolors=Violet}
\newcommand{\modelname}{{\sc RadNet}\xspace}
\newcommand{\dataset}{{\sc RadSet}\xspace}
\newcommand{\citet}[1]{\citeauthor{#1}~\shortcite{#1}}

\title{\modelname: Incident Prediction in Spatio-Temporal Road Graph Networks Using Traffic Forecasting}

\author{
Shreshth Tuli$^{1, 3}$\footnote{Indicates Equal Contribution.}\and
Matthew R. Wilkinson$^{2, 3 *}$\and
Chris Kettell$^{3}$\\
\affiliations
$^1$Imperial College London, UK\\
$^2$University of Bath, UK\\
$^3$TRL Software Limited, UK\\
\emails
s.tuli20@imperial.ac.uk,
mrw39@bath.ac.uk,
ckettell@trl.co.uk
}

\begin{document}

\maketitle

\begin{abstract}
Efficient and accurate incident prediction in spatio-temporal systems is critical to minimize service downtime and optimize performance. 
This work aims to utilize historic data to predict and diagnose incidents using spatio-temporal forecasting. We consider the specific use case of road traffic systems where incidents take the form of anomalous events, such as accidents or broken-down vehicles. To tackle this, we develop a neural model, called \modelname, which forecasts system parameters such as average vehicle speeds for a future timestep. As such systems largely follow daily or weekly periodicity, we compare \modelname's predictions against historical averages to label incidents. Unlike prior work, \modelname infers spatial and temporal trends in both permutations, finally combining the dense representations before forecasting. This facilitates informed inference and more accurate incident detection. Experiments with two publicly available and a new road traffic dataset demonstrate that the proposed model gives up to 8\% higher prediction F1 scores compared to the state-of-the-art methods. 
\end{abstract}

\section{Introduction}
\label{sec:intro}

Spatio-temporal forecasting is a critical problem for dynamic industrial environments such as transportation systems, supply chains and mobile computation systems~\cite{dcrnn,tascikaraoglu2018evaluation,yang2019short}. One such environment is the network of road traffic systems. Road networks act as systems that service mobile users where any incident could cause service downtimes~\cite{pan2013forecasting}. Here, road incidents take the form of anomalous events that deviate from expected trends, such as broken-down or parked vehicles. Thus, problem of incident prediction has gained attention specifically as a response to the growing populations, increasingly urbanized environments and a drive towards more sustainable transport. Additionally, diagnosing the specific root-cause of the incident (eg., the road where an accident has occurred) is crucial for informed remediation. 

Accurately forecasting and predicting incidents is challenging mainly due to the complex spatio-temporal dependencies in long-term forecasting. In particular, the increasing number of sensors and devices in contemporary traffic sensor platforms, calls for a more robust and scalable strategy. Moreover, the correlations between the traffic speeds and queue lengths of roads with those of upstream roads or junctions, requires a global analysis of the traffic network in lieu of a local one. Further, recurring incidents, such as accidents or illegal parking, give rise to non-stationarity, making long-term forecasting hard. To tackle this, prior work either uses knowledge-driven or data-driven methods, where the latter has been shown to outperform the former and hence considered as a \textit{de-facto} standard for traffic systems~\cite{dcrnn}. Data-driven methods leverage deep learning models that learn to predict future system states using historic data. State-of-the-art methods typically perform sequential inference of the spatial and temporal trends. Fixing a particular sequence in terms of capturing these trends assumes them to be independent. For instance, inferring spatial correlations using a graph-neural-network and sequentially inferring temporal trends using a recurrent neural network, makes the former oblivious to the analysis of the latter. This limits their performance as we demonstrate later.

\begin{figure}[t]
    \centering \setlength{\belowcaptionskip}{-15pt}
    \includegraphics[width=\linewidth]{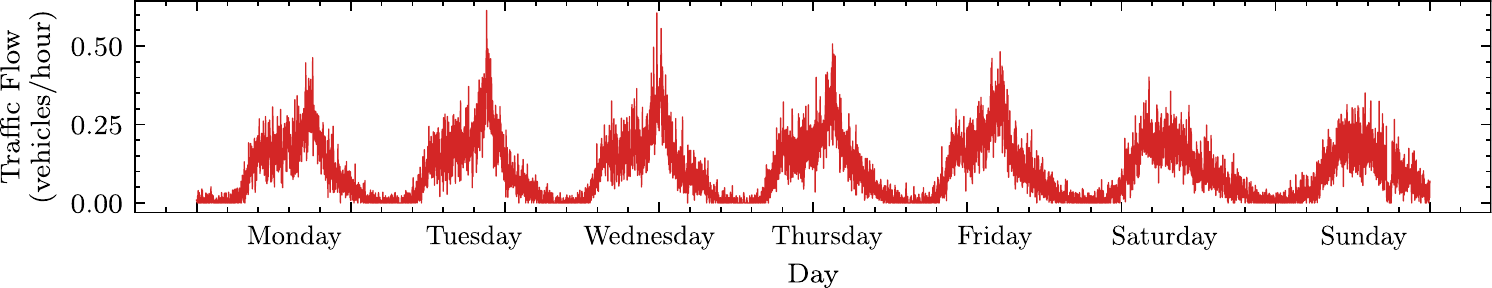} \vspace{-20pt}
    \caption{\footnotesize{Visualization of traffic flow for a single link in Radcliffe - UK, demonstrating the daily periodicity in traffic systems.}}
    \label{fig:motivation}
\end{figure}

In this paper, we study the important use case of utilizing a set of key parameters of a road traffic network to predict incidents/anomalies at a future timestep. Such parameters/features include flow rates, traffic speeds and queue lengths, characterizing the state of a road traffic system. We decompose the incident prediction problem into (i) forecasting the system characteristics at a future timestep using historic spatio-temporal traffic data, and (ii) comparing the forecasts against historic averages as a baseline to indicate the chance of an incident at that timestep. This is applicable in the domain of traffic systems as they largely conform to periodic (daily/weekly) trends (see Fig.~\ref{fig:motivation}). Thus, we propose a novel neural model, \underline{R}o\underline{a}d Incident Pre\underline{d}iction \underline{Net}work (\modelname), that addresses the previously highlighted limitation of fixed sequences in spatio-temporal models. It does this by explicitly inferring trends in both permutations and intelligently combining the outputs, facilitating an informed inference and consequently more accurate forecasts. Specifically, \modelname uses graph-attention (GAT) and Transformer encoders to infer spatial and temporal trends and combines the outputs using a custom weighted-skip connection. Experiments, using two publicly available datasets of vehicle speeds on highways, demonstrate that \modelname outperforms state-of-the-art benchmarks by increasing the incident prediction scores by up to 8\%. We also share a new dataset, called \dataset, constituting additional parameters such as queue lengths and congestion metrics for 24 non-highway roads (more prone to incidents than highways). \modelname outperforms benchmarks even on \dataset in in terms of incident prediction accuracy.

\vspace{-5pt}
\section{Related Work}
\label{sec:related_work}
Deep learning approaches have fast become the best practice in traffic forecasting, having outperformed more traditional statistical and machine learning methods such as Autoregressive Integrated Moving Average (ARIMA) and Support Vector Regression (SVR)~\cite{Lippi2013}. 

\noindent \textbf{Graph Oblivious Models.} 
A line of work aims to capture only temporal trends in traffic networks using either Gated Recurrent Units (GRUs) or Long Short Term Memory (LSTM) networks for traffic forecasting~\cite{lstm,lstm2}. To detect incidents, they have been extended with extreme-value-theory (EVT) based thresholding mechanisms. For instance,~\citet{lstm_ndt} propose an LSTM based forecasting model with a non-parametric dynamic error thresholding (NDT) strategy to label incidents using moving averages of the error sequence. However, the road sensor graph information is critical to utilize spatial correlations and neglecting this has been shown to adversely impact performance~\cite{stgcn}. Further, such models are slow to train and often unable to capture long-term trends. We include the LSTM based method by~\citet{lstm} as a benchmark in our experiments for completeness.

\noindent \textbf{Graph Convolution Models.} 
Several works utilize graph-convolution networks for traffic forecasting~\cite{gcn,chen2020multi}. In lieu of explicitly capturing the temporal trends in the data, such methods use a temporal sliding window of the sensor data for the last few timestamps to forecast characteristics of the traffic network at a future time. A recent method uses Graph Convolution Network (GCN)~\cite{gcn} to perform convolutions across neighboring nodes of the sensor graph and predict node features in a future interval. However, without explicitly modeling the temporal trends, such methods perform poorly in dynamic settings, typical for traffic networks. We include GCN as a benchmark in our experiments.

\noindent \textbf{Spatio-Temporal Graph Neural Networks.}
Most state-of-the-art methods for traffic forecasting and incident detection are based on spatio-temporal graph models~\cite{stgcn,graphwavenet_improved}. For instance, building upon the GCN model,~\citet{hagcn} propose a Hierarchical Attention-based Spatio-Temporal GCN model (HAGCN) for traffic forecasting using attention and graph convolutions being applied at different levels of the road network hierarchy. Another work, Diffusion Convolution RNN (DCRNN)~\cite{dcrnn}, utilizes diffusion graph convolutions and RNNs in tandem to model both spatial and temporal trends. Another method, called Deep Kalman Filtering Network (DKFN)~\cite{dkfn} uses GCN and RNNs to capture spatial trends and LSTMs to independently capture temporal trends of each node. It then combines the two using Kalman Filtering. More recently, Transformer blocks have been identified to provide accurate temporal models for forecasting, while additionally enabling parallelized computation for efficient training~\cite{Vaswani2017,tuli2022tranad}. A recent work, TSE-SC~\cite{tse_sc}, sequentially infers the input using a GCNs and then a Transformer encoder. Other works aim to map dependencies across nodes in the graph by explicitly learning a \textit{dependency matrix}. For instance, GTS~\cite{gts} first predicts a dependency matrix as dense vectors for each node and then applies recurrent graph convolutions for forecasting. Similarly, GraphWaveNet~\cite{graphwavenet} uses both graph and dilated causal convolutions for spatio-temporal inference with a self-learned dependency matrix of the graph. We use HAGCN, DCRNN, DKFN, TSE-SC, GTS and GraphWaveNet as benchmarks.


\begin{figure*}[t]
    \centering \setlength{\belowcaptionskip}{-15pt}
    \includegraphics[width=0.93\linewidth]{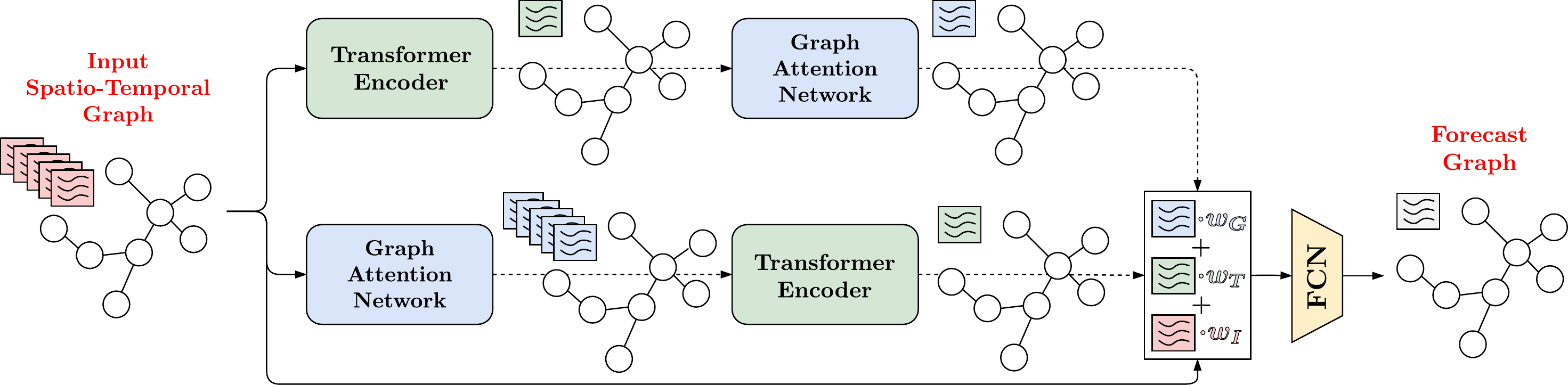} \vspace{-7pt}
    \caption{\footnotesize Neural architecture of the \modelname model. The model encodes the input spatio-temporal graph using two pathways: (i) temporal inference using a Transformer Encoder and spatial inference using a GAT, (ii) spatial inference using a GAT and temporal inference using a Transformer Encoder. \modelname then takes a convex combination of these two outputs and the last element of the input window to generate a forecast using a feed-forward network. The weights of the combination are generated using a feed-forward network not shown in the figure.} 
    \label{fig:nn}
\end{figure*}

\section{Methodology}

\subsection{Problem Formulation}
We consider the presence of a spatio-temporal trace, which is a timestamped sequence of datapoints. We first define a link as section of road with a single direction of travel and that connects two traffic junctions. Thus, a road that allows traffic to flow in opposite directions is represented as two links. Each link is represented by a node in the graph, with undirected edges used to represent connections to upstream and downstream links, \textit{i.e.}, links that share junctions (see Figure \ref{fig:radcliffe}). The graph of the road network is denoted by $G = (V, E)$ such that $V$ denotes the set of $N$ links and $E$ denotes the set of edges across links, both assumed to be static. We denote the spatio-temporal trace by $\mathcal{T} = \{X^{(1)}, \ldots, X^{(T)}\}$ with size $T$. Here, at each timestep $t$, the graph $G$ has a feature matrix denoted by $X^{(t)}$ that represents the collection of data features for all links, such as average vehicular speed, flow rate and queue lengths. We assume the presence of $D$ features for each link at each timestep. Thus, $X^{(t)} \in \mathds{R}^{N \times D}$. Moreover, a timestep $t \in \{1, \ldots, T\}$ is a discretized temporal interval, each of a duration $\Delta$, over which the data features are aggregated to form $X^{(t)}$. 
At each timestep $t$, we consider a local contextual window of length $K$ as
\begin{equation*}
    W^{(t)} = X^{(t-K):t},
\end{equation*}
where we use replication padding for $t < K$. We also refer $W^{(t)}$ as the \textit{state} of the system. This work aims to learn a function $f_\theta(.)$ parameterized by the set $\theta$ to forecast graph features after $H$ steps. Thus,
\begin{equation*}
    \hat{X}^{(t+H)} = f_\theta(W^{(t)}, G).
\end{equation*}
We define \textit{time horizon} as the look-ahead time we want to forecast to, which we use to calculate $H$. For instance, with an interval duration of $\Delta = $ 5 minutes and time horizon of 15 minutes, $H = 3$ intervals. 

To generate incident labels, we leverage the daily and weekly periodicity of road traffic networks~\cite{scoot}. For each timestep $t$, we denote the weekday for $t$ by $d(t)$ and the 24-hour clock time by $c(t)$. Now, for each timestamp $t$ and dataset $\mathcal{T}$, we define a baseline of historical averages that aggregates the feature matrices in the dataset that belong to the same weekday as $t$ and differ in terms of clock-time by $\leq \Delta$. Thus,
\begin{equation*}
    B^{(t)} = \mathrm{Mean} \{X^{(u)} \mid d(t) = d(u) \wedge |c(t) - c(u)| \leq \Delta \}.
\end{equation*}
We define residuals for predicted and true feature matrices as
\begin{align}
\begin{split}
\label{eq:residuals}
    \hat{S}^{(t+H)} = \| B^{(t+H)} - \hat{X}^{(t+H)} \|,\\
    S^{(t+H)} = \| B^{(t+H)} - X^{(t+H)} \|.
\end{split}
\end{align}
We use a dynamic thresholding policy described later to generate a threshold at each timestep $t$, denoted by $\phi(t)$. Our incident label is then defined as a binary value
\begin{align}
\begin{split}
\label{eq:incident_labels}
    \hat{Y}^{(t+H)} &= \mathds{1}(\hat{S}^{(t+H)} \geq \phi(t+H)),\\
    Y^{(t+H)} &= \mathds{1}(S^{(t+H)} \geq \phi(t+H)).
\end{split}
\end{align}
This enables us to generate ground-truth incident labels $Y^{(t)}$ without explicitly using expert labelling. We denote train and test datasets by $\mathcal{T}_{train}$ and $\mathcal{T}_{test}$, both formed by partitioning $\mathcal{T}$. Our objective is to ensure that the predicted incident labels $\hat{Y}^{(t)}$ resemble closely the true labels $Y^{(t)}$ for $\mathcal{T}_{test}$ given $f_\theta(.)$ is trained using $\mathcal{T}_{train}$.

\subsection{\modelname Model}
The function $f_\theta(.)$ is dubbed as \modelname in our work and its architecture is summarized in Fig.~\ref{fig:nn}. For a graph $G$ of a road network, at each timestep $t$, we encode the input window $W^{(t)}$ to perform both permutations of spatial and temporal inferences as described in Section~\ref{sec:intro}. For spatial inference, we use graph attention networks (GAT)~\cite{Vel2017} and for temporal inference we use transformer blocks~\cite{Vaswani2017}.

\noindent \textbf{Spatial-Inference.} For an input feature matrix $X^{(t)}$, we denote the feature of each node $i \in \{1, \ldots, N\}$ as $h^{(t)}_i$. We denote the \textit{neighborhood} of node $i$ in the graph by $\mathcal{N}_i$. We compute attention coefficients using GAT weights $\theta^G_1$ as
\begin{align*}
\begin{split}
    \alpha_{ij} &= \mathrm{softmax}_j (e_{ij}) = \frac{\mathrm{exp}(e_{ij})}{ \sum_{l \in \mathcal{N}_i} \mathrm{exp}(e_{il})} , \text{ where}\\
    e_{ij} &= \mathrm{LeakyReLU} (\mathrm{FeedForward}(\cdot [\theta^G_1 \cdot h_i, \theta^G_1 \cdot h_j])).
\end{split}
\end{align*}
For given input feature matrix $X^{(t)}$, we perform multi-head graph attention with $M$ heads as 
\begin{equation*}
    h'_i = \mathrm{Concat}\Big(\Big\{\mathrm{sigmoid} \Big(\sum_{j \in \mathcal{N}_i} \alpha^m_{ij}\ \theta^G_1\ h_j \Big)\Big\}_{i = 1}^M \Big).
\end{equation*}
We aggregate the outputs of the multiple heads in the final layer to ensure same feature size as the input~\cite{Vel2017}. The stacked matrix of $h'_i \forall i$ is denoted by 
\begin{equation}
    X^{(t)}_1 = \mathrm{GAT}(X^{(t)}).
\end{equation}

\noindent \textbf{Temporal Inference.} For temporal inference over a sequence, we use a Transformer model. We define multi-head attention using the scaled-dot product attention from~\citet{Vaswani2017} for an input $W^{(t)}$ after position-encoding $W^{(t)}_p$:
\begin{align*}
\small
\begin{split}
    \mathrm{MHA(I, I, I)} &= \mathrm{Concat}(I_1, \ldots, I_O) I, \text{ where}\\
    I_q &= \mathrm{softmax}\Big(\frac{I W^Q_q \cdot (I W^K_q)^T}{\sqrt{D}} I W^V_q \Big),\\
    W^{(t)}_{11} &= \mathrm{LayerNorm}(W^{(t)}_p + \mathrm{MHA}(W^{(t)}_p, W^{(t)}_p, W^{(t)}_p)),\\
    W^{(t)}_{12} &= \mathrm{LayerNorm}(W^{(t)}_{11} + \mathrm{FeedForward}(W^{(t)}_p)),
\end{split}
\end{align*}
where $\mathrm{LayerNorm}$ represents the layer-normalization operation. We also give masked output $W^{(t+K)}$ after position encoding as $W^{(t+k)}_p$ to generate $W^{(t)}_1$ as
\begin{align*}
\begin{split}
\label{eq:window_encoder}
    W^{(t+k)}_{11} &= \mathrm{Mask}(\mathrm{MHA}(W^{(t+k)}_p, W^{(t+k)}_p, W^{(t+k)}_p)), \\
    W^{(t+k)}_{12} &= \mathrm{LayerNorm}(W^{(t+k)}_p + W^{(t+k)}_{11}),\\
    W^{(t)}_1 &= \mathrm{LayerNorm}(I_2^2 + \mathrm{MHA}(W^{(t)}_{12}, W^{(t)}_{12}, W^{(t+k)}_{12})).
\end{split}
\end{align*}
The above steps are represented as 
\begin{equation}
    W^{(t)}_1 = \mathrm{Transformer}(W^{(t)}).
\end{equation}

\begin{figure*}
    \centering \setlength{\belowcaptionskip}{-10pt}
    \includegraphics[width=\linewidth]{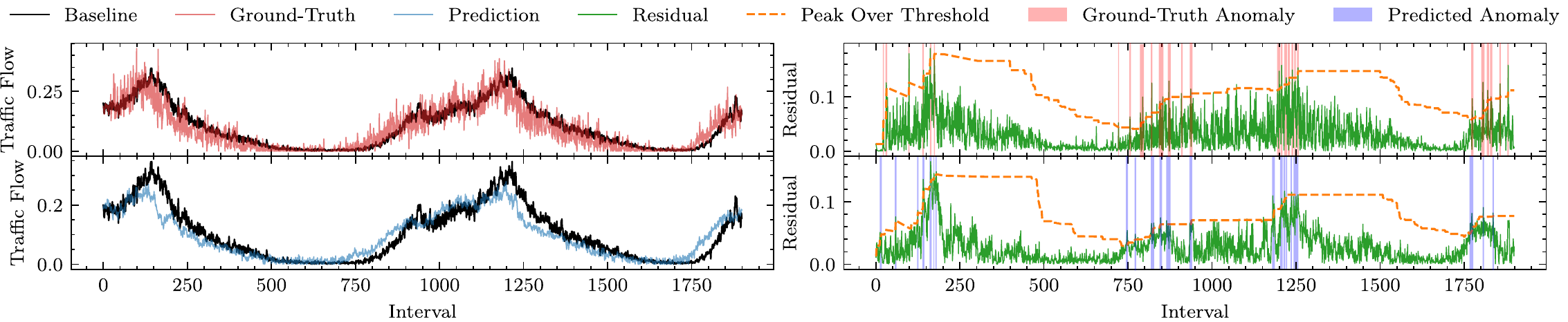}
    \caption{Visualization of (left) baseline, ground-truth and predicted time-series, (right) residual, Peak-Over-Threshold value and anomaly labels. The visualizations have been shown for \dataset dataset for a sample link and the \textit{flow rate} feature. Traffic flow is normalized \textit{vehicles per hour}, whereas each interval is of the duration of \textit{5 minutes}. Predictions generated using \modelname.}
    \label{fig:visualization}
\end{figure*}

\noindent \textbf{Multi-Inference Model.} For an input window $W^{(t)}$, we perform spatio-temporal inference as
\begin{align}
\label{eq:spatio_temporal}
\begin{split}
W^{(t)}_G &= \mathrm{GAT}(W^{(t)}),\\
W^{(t)}_1 &= \mathrm{GAT}(W^{(t)}_G).
\end{split}
\end{align}
where we perform GAT operations in a factored style over each feature matrix in $W^{(t)} = X^{(t-K):t}$ to generate $W^{(t)}_G = X^{(t-K):t}_1$. We also perform temporo-spatial inference using Transformer and GAT in the reverse order
\begin{align}
\label{eq:temporo_spatial}
\begin{split}
W^{(t)}_T &= \mathrm{Transformer}(W^{(t)}),\\
W^{(t)}_2 &= \mathrm{GAT}(W^{(t)}_T).
\end{split}
\end{align}
To combine the outputs of the above two inference structures, we use a feed-forward model to generate attention weights as
\begin{align*}
w &= \mathrm{FeedForward}(X^{(t)}),\\
\hat{X}^{(t+H)} &= \mathrm{FeedForward}(w^T \cdot [W^{(t)}_1, W^{(t)}_2, X^{(t)}]).
\end{align*}
The weights used to take convex combination of the spatio-temporal, temporo-spatial and the input facilitate assigning dynamic importance coefficients to the three encodings. The final forecast then becomes $\hat{X}^{(t+H)} = f_\theta(W^{(t)})$ where the $\theta$ parameters include weights of the two GAT, two Transformer blocks and feed-forward networks. Using a training dataset $\mathcal{T}_{train}$ the loss for each timestep $t$ defined as
\begin{equation*}
    L(t) = \| X^{(t+H)} - \hat{X}^{(t+H)} \|.
\end{equation*}
For a training dataset $\mathcal{T}_{train}$, use the loss function  $\sum_{t = 0}^{|\mathcal{T}_{train}| - H} L(t)$ to tune the $\theta$ parameters. 

\begin{figure}
    \centering \setlength{\belowcaptionskip}{-10pt}
    \includegraphics[width=\linewidth]{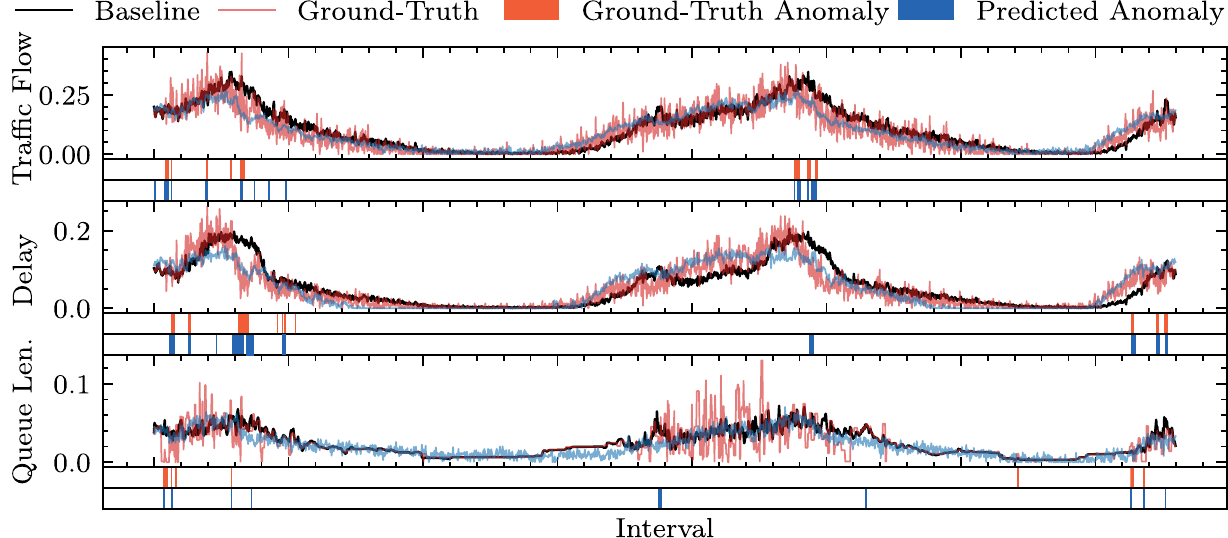}
    \caption{Visualization of baseline, ground-truth and predicted time-series for normalized traffic flow (vehicles per hour), delay (seconds), queue length and anomaly labels in \dataset dataset using \modelname model.}
    \label{fig:vis_all}
\end{figure}

\noindent \textbf{Incident Prediction.} Once we have a trained model, we use the predictions to generate the residuals and labels using eqs.~\ref{eq:residuals} and~\ref{eq:incident_labels}. The residuals give us an indication of how far the predicted feature matrix is compared to the historic averages. A high residual indicates an anomalous event and hence we raise an incident flag when this residual value is above a threshold. As is common in prior work~\cite{omnianomaly,lstm_ndt}, we use the Peak Over Threshold (POT)~\cite{siffer2017anomaly} method to choose the threshold automatically and dynamically. In essence, this is a statistical method that uses ``extreme value theory'' to fit the data distribution with a Generalized Pareto Distribution and identify appropriate values at risk to dynamically determine threshold values. Fig.~\ref{fig:visualization} visualizes the \textit{traffic-flow} feature in \dataset for a sample link. The plots on the left show baseline, ground-truth time-series and predicted series. The plots on the right show residuals and incident labels, both for predicted and ground-truth series. As shown, whenever the residual values exceed POT thresholds, the incident label is one and zero otherwise. The residual and label generation can then be performed for each link in the graph as shown in Fig.~\ref{fig:vis_all}.

\section{Experiments}
\label{sec:experiments}

We compare \modelname with state-of-the-art models for spatio-temporal forecasting or incident detection, including DCRNN~\cite{dcrnn}, GTS~\cite{gts}, GCN~\cite{gcn}, HAGCN~\cite{hagcn}, LSTM~\cite{lstm}, TSE-SC~\cite{tse_sc}, DKFN~\cite{dkfn} and GraphWaveNet~\cite{graphwavenet}. For more details refer to Section~\ref{sec:related_work}.\footnote{We use publicly available code sources for most baselines. DCRNN~\url{https://github.com/chnsh/DCRNN_PyTorch}, GTS~\url{https://github.com/chaoshangcs/GTS} and DKFN~\url{https://github.com/Fanglanc/DKFN}. We use the improved version of GraphWaveNet by~\citet{graphwavenet_improved} available at \url{https://github.com/sshleifer/Graph-WaveNet}. Other models were re-implemented by us.} We tune the hyperparameters of all methods as per grid-search. We train all models using the PyTorch-1.7.1~\cite{paszke2019pytorch} library. All model training and experiments were performed on an Amazon EC2 G4 instance with 48 core CPU, 2x Tesla T4 GPUs and 192 GB RAM.

To train \modelname, we divide the time-series data using the 5-fold cross-validation process. To avoid over-fitting, we use the early-stopping criteria using the value of the loss function on the validation set. We use the AdamW~\cite{kingma2014adam} optimizer to train our model. See Appendices~\ref{sec:arch} and~\ref{sec:hyperparam} for the specific hyperparameter values used in our experiments. The only dataset-specific hyperparameter is the number of heads in multi-head graph-attention, which was kept to be the same as the dimension size of the dataset. Other assignments for this hyperparameter give similar trends. 

\subsection{Datasets}

\begin{table}[] 
    \centering \setlength{\belowcaptionskip}{-10pt}
    \resizebox{\linewidth}{!}{
\begin{tabular}{@{}lcccc@{}}
\toprule
Dataset & Timesteps & Num. Nodes & Num. Edges & Features\tabularnewline
\midrule
METR-LA & 34272 & 207 & 2833 & 1\tabularnewline
PEMS & 52116 & 325 & 4483 & 1\tabularnewline
\dataset & 209739 & 24 & 166 & 7\tabularnewline
\bottomrule
\end{tabular}}
    \caption{\footnotesize Dataset Characteristics.}
    \label{tab:dset_info}
\end{table}

We use three datasets to verify the \modelname model, with characteristics for each shown in Table \ref{tab:dset_info}. The first two (METR-LA and PEMS) are open-source traffic network datasets by~\citet{dcrnn}, commonly used in prior work. 

\noindent \textbf{METR-LA} 
is a dataset that comes from 207 loop detector sensors that record the vehicular speeds along the highways in Los Angeles County, USA. This dataset contains only the speed parameter for each link in the area and spans four months (March 1st 2012 to June 30th 2012) with a data collection interval of 5 minutes. 

\noindent \textbf{PEMS} 
(Performance Measurement System) by California Transport Agencies provides data from 325 sensors covering the San Francisco Bay Area, USA. Similar to METR-LA, this also contains only the speed information of vehicles and spans six months (Jan 1st 2017 to May 31st 2017) with intervals of 5 minute duration.

\noindent {\bfseries\scshape RadSet} 
is a new spatio-temporal dataset for traffic analysis that we present as part of this work. The above two traffic datasets cover highway networks in the United States. Performance of models trained on these data sets may not translate to European road networks. Moreover, the above two datasets capture only speed data for highways. There is a dearth of datasets that capture data for non-highway road networks where the traffic lights and illegal parking may give rise to incidents not as common in highways. Further, the above two datasets lack other parameters of a road network, such as queue lengths at traffic signals. Thus, we present for the first time, the \dataset data set collected from the Open Data Portal of the TRL SCOOT (Split Cycle and Offset Optimisation Technique) UTC (Urban Traffic Control) system~\cite{scoot}. \dataset covers a network of 24 detectors in Radcliffe, Greater Manchester, UK. Unlike the other two data sets, where only the aggregated speed readings are recorded, \dataset provides 7 features at each detector including flow rate (vehicles per hour), congestion index and queue lengths. The collection interval is the traffic signal cycle time. Further details on the \dataset dataset presented in Appendix~\ref{sec:datset_details}.  

\begin{table}[t]
    \centering \setlength{\belowcaptionskip}{-5pt}
    \resizebox{\linewidth}{!}{
\begin{tabular}{@{}lcccccccc@{}}
\toprule 
Time & \multirow{2}{*}{Model} & \multicolumn{7}{c}{\dataset}\tabularnewline
\cmidrule{3-9}
Horizon &  & P & R & F1 & H@1 & H@1.5 & N@1 & N@1.5\tabularnewline
\midrule
\multirow{9}{*}{5 min} & DCRNN & 0.592 & \textbf{0.990} & 0.741 & 0.489 & 0.568 & 0.481 & 0.549\tabularnewline
 & GTS & 0.880 & 0.961 & 0.919 & 0.653 & 0.764 & 0.652 & 0.749\tabularnewline
 & GCN & 0.854 & 0.925 & 0.888 & 0.630 & 0.732 & 0.623 & 0.713\tabularnewline
 & HAGCN & 0.858 & 0.960 & 0.906 & 0.632 & 0.734 & 0.623 & 0.715\tabularnewline
 & LSTM & 0.441 & 0.473 & 0.456 & 0.457 & 0.521 & 0.450 & 0.504\tabularnewline
 & TSE-SC & 0.865 & 0.944 & 0.902 & 0.626 & 0.733 & 0.629 & 0.719\tabularnewline
 & DKFN & 0.498 & 0.196 & 0.281 & 0.487 & 0.566 & 0.492 & 0.560\tabularnewline
 & GraphWaveNet & 0.637 & 0.182 & 0.283 & 0.538 & 0.627 & 0.542 & 0.619\tabularnewline
 & \modelname & \textbf{0.925} & 0.935 & \textbf{0.930} & \textbf{0.689} & \textbf{0.750} & \textbf{0.685} & \textbf{0.740}\tabularnewline
\midrule
\multirow{9}{*}{15 min} & DCRNN & 0.572 & \textbf{0.992} & 0.726 & 0.482 & 0.558 & 0.474 & 0.539\tabularnewline
 & GTS & 0.771 & 0.973 & 0.860 & 0.567 & 0.676 & 0.567 & 0.667\tabularnewline
 & GCN & 0.822 & 0.933 & 0.874 & 0.596 & 0.703 & 0.592 & 0.683\tabularnewline
 & HAGCN & 0.823 & 0.963 & 0.888 & 0.597 & 0.714 & 0.594 & 0.692\tabularnewline
 & LSTM & 0.436 & 0.505 & 0.468 & 0.453 & 0.514 & 0.447 & 0.500\tabularnewline
 & TSE-SC & 0.750 & 0.968 & 0.845 & 0.566 & 0.674 & 0.553 & 0.648\tabularnewline
 & DKFN & 0.492 & 0.292 & 0.366 & 0.492 & 0.572 & 0.489 & 0.557\tabularnewline
 & GraphWaveNet & 0.575 & 0.252 & 0.350 & 0.530 & 0.621 & 0.533 & 0.611\tabularnewline
 & \modelname & \textbf{0.889} & 0.949 & \textbf{0.918} & \textbf{0.659} & \textbf{0.762} & \textbf{0.653} & \textbf{0.741}\tabularnewline
\midrule
\multirow{9}{*}{30 min} & DCRNN & 0.556 & \textbf{0.993} & 0.713 & 0.475 & 0.549 & 0.468 & 0.531\tabularnewline
 & GTS & 0.704 & 0.980 & 0.819 & 0.536 & 0.640 & 0.528 & 0.616\tabularnewline
 & GCN & 0.802 & 0.967 & 0.876 & 0.592 & 0.699 & 0.579 & 0.674\tabularnewline
 & HAGCN & 0.804 & 0.955 & 0.873 & 0.590 & 0.696 & 0.580 & 0.675\tabularnewline
 & LSTM & 0.432 & 0.531 & 0.477 & 0.452 & 0.513 & 0.445 & 0.498\tabularnewline
 & TSE-SC & 0.678 & 0.983 & 0.802 & 0.527 & 0.620 & 0.513 & 0.595\tabularnewline
 & DKFN & 0.502 & 0.352 & 0.414 & 0.493 & 0.574 & 0.492 & 0.559\tabularnewline
 & GraphWaveNet & 0.578 & 0.356 & 0.440 & 0.521 & 0.608 & 0.521 & 0.596\tabularnewline
 & \modelname & \textbf{0.857} & 0.950 & \textbf{0.901} & \textbf{0.633} & \textbf{0.742} & \textbf{0.626} & \textbf{0.722}\tabularnewline
\midrule
\multirow{9}{*}{60 min} & DCRNN & 0.540 & \textbf{0.994} & 0.700 & 0.469 & 0.539 & 0.463 & 0.523\tabularnewline
 & GTS & 0.698 & 0.981 & 0.816 & 0.531 & 0.635 & 0.526 & 0.614\tabularnewline
 & GCN & 0.793 & 0.968 & 0.872 & 0.579 & 0.685 & 0.576 & 0.671\tabularnewline
 & HAGCN & 0.780 & 0.943 & 0.854 & 0.575 & 0.680 & 0.569 & 0.662\tabularnewline
 & LSTM & 0.432 & 0.564 & 0.489 & 0.448 & 0.510 & 0.443 & 0.495\tabularnewline
 & TSE-SC & 0.630 & 0.984 & 0.768 & 0.508 & 0.590 & 0.494 & 0.567\tabularnewline
 & DKFN & 0.473 & 0.405 & 0.436 & 0.477 & 0.551 & 0.474 & 0.538\tabularnewline
 & GraphWaveNet & 0.528 & 0.370 & 0.435 & 0.515 & 0.591 & 0.502 & 0.571\tabularnewline
 & \modelname & \textbf{0.827} & 0.961 & \textbf{0.889} & \textbf{0.608} & \textbf{0.716} & \textbf{0.599} & \textbf{0.694}\tabularnewline
\bottomrule
\end{tabular}}
    \caption{\footnotesize Performance comparison of \modelname with the state-of-the-art benchmarks on the \dataset dataset. H and N signify HitRate and NDCG incident diagnosis metrics, respectively.}
    \label{tab:resutls_scoots}
\end{table}

\subsection{Evaluation Metrics}
\noindent \textbf{Incident Detection.} 
To evaluate the efficacy of the incident predictions, we use precision (P), recall (R) and F1 score. In a deployment scenario where the flagged incidents are acted upon through operators, we aim to minimize both false-negative and false-positive. Minimizing the former is crucial to eschew adverse impacts of incidents on road traffic, whereas reducing the latter is important to avoid excessive cost implications of unnecessary remediation steps. Thus, we consider the F1 score as the central metric and use it for our hyperparameter search (see Appendix~\ref{sec:hyperparam}).

\noindent \textbf{Incident Diagnosis.} 
We also measure the incident diagnosis performance, such that we could identify the links at which incidents occur, rather than just determining there is an incident somewhere in the entire network. To measure the diagnosis performance, we use HitRate@P\%~\cite{hitrate} and NDCG@P\% (normalized discounted cumulative gain)~\cite{ndcg} where P\% is 100 and 150 for both metrics as seen in prior work~\cite{tuli2022tranad,hitpvalue}. Here, these assess how many of the ground-truth links have been included in the top predicted incident candidates. To give an example, if at a given timestamp $t$, there are 8 links with incidents, then P\% = 100 would consider the top 8 predicted links and P\% = 150 the top 12. 


\begin{table*}[t]
    \centering \setlength{\belowcaptionskip}{-6pt}
    \resizebox{\linewidth}{!}{
\begin{tabular}{@{}lccccccccccccccc@{}}
\toprule 
Time & \multirow{2}{*}{Model} & \multicolumn{7}{c}{METR-LA} & \multicolumn{7}{c}{PEMS}\tabularnewline
\cmidrule{3-16}
Horizon &  & P & R & F1 & H@1 & H@1.5 & N@1 & N@1.5 & P & R & F1 & H@1 & H@1.5 & N@1 & N@1.5\tabularnewline
\midrule
\multirow{9}{*}{5 min} & DCRNN & 0.372 & 0.334 & 0.352 & 0.399 & 0.695 & 0.393 & 0.652 & 0.523 & 0.428 & 0.471 & 0.532 & 0.815 & 0.532 & 0.779\tabularnewline
 & GTS & 0.369 & 0.332 & 0.349 & 0.396 & 0.693 & 0.391 & 0.651 & 0.513 & 0.417 & 0.460 & 0.527 & 0.815 & 0.524 & 0.775\tabularnewline
 & GCN & 0.535 & 0.484 & 0.508 & 0.530 & 0.758 & 0.530 & 0.732 & 0.566 & 0.461 & 0.508 & 0.563 & 0.826 & 0.563 & 0.794\tabularnewline
 & HAGCN & 0.545 & 0.490 & 0.516 & 0.538 & 0.769 & 0.538 & 0.737 & 0.545 & 0.444 & 0.490 & 0.549 & 0.822 & 0.548 & 0.787\tabularnewline
 & LSTM & 0.564 & 0.506 & 0.533 & 0.551 & 0.769 & 0.553 & 0.745 & 0.647 & 0.526 & 0.580 & 0.616 & 0.851 & 0.620 & 0.823\tabularnewline
 & TSE-SC & 0.551 & 0.498 & 0.523 & 0.543 & 0.768 & 0.543 & 0.739 & 0.564 & 0.460 & 0.507 & 0.560 & 0.828 & 0.561 & 0.794\tabularnewline
 & DKFN & 0.638 & 0.575 & 0.605 & 0.612 & 0.801 & 0.616 & 0.782 & 0.660 & 0.538 & 0.593 & 0.625 & 0.855 & 0.630 & 0.828\tabularnewline
 & GraphWaveNet & \textbf{0.875} & 0.459 & 0.603 & 0.637 & 0.817 & 0.655 & 0.812 & \textbf{0.671} & 0.567 & 0.614 & \textbf{0.644} & 0.852 & 0.612 & 0.819\tabularnewline
 & \modelname & 0.679 & \textbf{0.677} & \textbf{0.678} & \textbf{0.676} & \textbf{0.840} & \textbf{0.676} & \textbf{0.816} & 0.620 & \textbf{0.614} & \textbf{0.617} & 0.619 & \textbf{0.853} & \textbf{0.619} & \textbf{0.821}\tabularnewline
\midrule
\multirow{9}{*}{15 min} & DCRNN & 0.363 & 0.309 & 0.334 & 0.398 & 0.695 & 0.394 & 0.652 & 0.522 & 0.402 & 0.454 & 0.537 & 0.818 & 0.533 & 0.779\tabularnewline
 & GTS & 0.349 & 0.297 & 0.321 & 0.390 & 0.691 & 0.384 & 0.646 & 0.519 & 0.401 & 0.452 & 0.534 & 0.815 & 0.531 & 0.778\tabularnewline
 & GCN & 0.527 & 0.450 & 0.486 & 0.524 & 0.760 & 0.521 & 0.726 & 0.584 & 0.451 & 0.509 & 0.572 & 0.831 & 0.573 & 0.800\tabularnewline
 & HAGCN & 0.533 & 0.454 & 0.490 & 0.524 & 0.757 & 0.525 & 0.729 & 0.561 & 0.432 & 0.488 & 0.559 & 0.824 & 0.558 & 0.792\tabularnewline
 & LSTM & 0.556 & 0.472 & 0.510 & 0.540 & 0.760 & 0.542 & 0.739 & 0.657 & 0.505 & 0.571 & 0.615 & 0.851 & 0.621 & 0.824\tabularnewline
 & TSE-SC & 0.541 & 0.462 & 0.499 & 0.532 & 0.765 & 0.532 & 0.733 & 0.562 & 0.433 & 0.489 & 0.560 & 0.825 & 0.559 & 0.793\tabularnewline
 & DKFN & 0.635 & 0.542 & 0.585 & 0.599 & 0.788 & 0.604 & 0.776 & 0.668 & 0.514 & 0.581 & 0.622 & 0.854 & 0.628 & 0.827\tabularnewline
 & GraphWaveNet & \textbf{0.856} & 0.408 & 0.553 & 0.616 & 0.810 & 0.632 & 0.799 & \textbf{0.671} & 0.514 & 0.582 & 0.6225 & 0.852 & 0.621 & 0.823\tabularnewline
 & \modelname & 0.679 & \textbf{0.644} & \textbf{0.661} & \textbf{0.662} & \textbf{0.833} & \textbf{0.664} & \textbf{0.809} & 0.630 & \textbf{0.596} & \textbf{0.613} & \textbf{0.623} & \textbf{0.855} & \textbf{0.624} & \textbf{0.824}\tabularnewline
\midrule
\multirow{9}{*}{30 min} & DCRNN & 0.326 & 0.261 & 0.290 & 0.381 & 0.687 & 0.377 & 0.641 & 0.522 & 0.377 & 0.438 & 0.538 & 0.819 & 0.534 & 0.780\tabularnewline
 & GTS & 0.384 & 0.308 & 0.342 & 0.421 & 0.707 & 0.418 & 0.665 & 0.524 & 0.379 & 0.440 & 0.539 & 0.817 & 0.536 & 0.781\tabularnewline
 & GCN & 0.521 & 0.423 & 0.467 & 0.517 & 0.750 & 0.515 & 0.723 & 0.568 & 0.412 & 0.478 & 0.562 & 0.827 & 0.562 & 0.794\tabularnewline
 & HAGCN & 0.525 & 0.421 & 0.467 & 0.517 & 0.752 & 0.518 & 0.725 & 0.562 & 0.408 & 0.472 & 0.560 & 0.827 & 0.558 & 0.792\tabularnewline
 & LSTM & 0.546 & 0.436 & 0.485 & 0.527 & 0.757 & 0.532 & 0.733 & 0.662 & 0.478 & 0.555 & 0.613 & 0.850 & 0.618 & 0.823\tabularnewline
 & TSE-SC & 0.529 & 0.426 & 0.472 & 0.521 & 0.753 & 0.521 & 0.726 & 0.567 & 0.412 & 0.477 & 0.561 & 0.826 & 0.562 & 0.794\tabularnewline
 & DKFN & 0.627 & 0.504 & 0.559 & 0.584 & 0.784 & 0.590 & 0.768 & 0.676 & 0.491 & 0.569 & 0.620 & 0.853 & 0.627 & 0.827\tabularnewline
 & GraphWaveNet & \textbf{0.795} & 0.337 & 0.474 & 0.585 & 0.794 & 0.597 & 0.777 & \textbf{0.664} & 0.483 & 0.559 & 0.611 & 0.847 & \textbf{0.620} & 0.814\tabularnewline
 & \modelname & 0.681 & \textbf{0.613} & \textbf{0.645} & \textbf{0.648} & \textbf{0.820} & \textbf{0.652} & \textbf{0.803} & 0.624 & \textbf{0.562} & \textbf{0.591} & \textbf{0.612} & \textbf{0.851} & 0.613 & \textbf{0.818}\tabularnewline
\midrule
\multirow{9}{*}{60 min} & DCRNN & 0.340 & 0.257 & 0.292 & 0.402 & 0.701 & 0.396 & 0.652 & 0.539 & 0.369 & 0.438 & 0.548 & 0.823 & 0.545 & 0.785\tabularnewline
 & GTS & 0.363 & 0.274 & 0.312 & 0.411 & 0.706 & 0.411 & 0.661 & 0.541 & 0.369 & 0.439 & 0.548 & 0.820 & 0.546 & 0.786\tabularnewline
 & GCN & 0.452 & 0.342 & 0.390 & 0.474 & 0.735 & 0.469 & 0.695 & 0.580 & 0.394 & 0.469 & 0.568 & 0.828 & 0.568 & 0.797\tabularnewline
 & HAGCN & 0.519 & 0.392 & 0.447 & 0.516 & 0.758 & 0.512 & 0.721 & 0.579 & 0.396 & 0.470 & 0.568 & 0.831 & 0.567 & 0.797\tabularnewline
 & LSTM & 0.538 & 0.403 & 0.461 & 0.521 & 0.758 & 0.524 & 0.728 & 0.673 & 0.457 & 0.545 & 0.611 & 0.849 & 0.619 & \textbf{0.824}\tabularnewline
 & TSE-SC & 0.526 & 0.399 & 0.454 & 0.516 & 0.759 & 0.517 & 0.724 & 0.572 & 0.390 & 0.464 & 0.564 & 0.827 & 0.564 & 0.795\tabularnewline
 & DKFN & 0.625 & 0.470 & 0.536 & 0.580 & 0.787 & 0.581 & 0.762 & 0.676 & 0.460 & 0.547 & 0.610 & \textbf{0.850} & 0.620 & \textbf{0.824}\tabularnewline
 & GraphWaveNet & \textbf{0.740} & 0.276 & 0.402 & 0.562 & 0.781 & 0.569 & 0.760 & \textbf{0.656} & 0.445 & 0.531 & 0.602 & 0.843 & 0.610 & 0.819\tabularnewline
 & \modelname & 0.671 & \textbf{0.573} & \textbf{0.618} & \textbf{0.628} & \textbf{0.808} & \textbf{0.633} & \textbf{0.792} & 0.631 & \textbf{0.542} & \textbf{0.583} & \textbf{0.611} & \textbf{0.850} & \textbf{0.613} & 0.819\tabularnewline
\bottomrule
\end{tabular}}
    \caption{Performance of \modelname and the state-of-the-art benchmarks on the METR-LA and PEMS datasets.}
    \label{tab:results_metrla_pems}
\end{table*}

\subsection{Results}

\noindent \textbf{Incident Detection.} Tables~\ref{tab:resutls_scoots} and~\ref{tab:results_metrla_pems} presents results for \modelname and benchmark models for the three datasets. We consider four values of the time horizon (for $H$): 5, 15, 30 and 60 minutes. For the \dataset dataset, \modelname outperforms all models across all metrics. \modelname gives an average F1 score of 0.909, which is 3.6\% higher than the second-best average score of the GCN model (0.877). For the METR-LA dataset, the \modelname model gives an average F1 score of 0.651, which is 14.0\% higher than the next best average F1 of the DKFN model (0.571). Similarly, \modelname also gives the highest average F1 score of 0.601 on the PEMS dataset, 4.7\% higher than the highest average F1 of GraphWaveNet of 0.574. We also observe that the gains due to the \modelname model increase for higher time horizons. For instance, for the METR-LA dataset, the performance gain in F1 score compared to the DKFN model increases from 12.1\% for a 5-minute horizon to 15.3\% for a 60-minute horizon. We hypothesize this is due to the decoupled spatio-temporal and temporo-spatial inference with skip connection, facilitating the downstream feed-forward predictors to use these encoding independently and avoiding the vanishing gradient problem in benchmark models~\cite{wang2018predrnn++}.

\noindent \textbf{Incident Diagnosis.} In terms of the diagnosis performance, which measures the ability of the model to identify the anomalous links, \modelname outperforms all benchmark models for the three datsets. For instance, for the \dataset dataset, the diagnosis scores improve by 5.39\%-8.09\% compared to the best benchmark model, \textit{i.e.}, GCN. Similarly, for the METR-LA and PEMS datasets, the improvements lie in the range 2.28\%-8.83\% compared to GraphWaveNet. Improvements in incident diagnosis are facilitated by the multi-head attentions in the GAT and Transformer blocks, enabling \modelname to attend to multiple links simultaneously, making it more suitable for inter-link incidents. This is observed and explained by the \dataset dataset, where incidents in one link can lead to a chain of events causing incidents in neighboring links as well. \modelname is able to accurately identify links where incidents occur, making it ideal for informed incident remediation. 
We present ablation analysis, model size comparison and further explorations in Appendix~\ref{sec:additional_results}.
        
\section{Conclusions}
This work presents a novel neural architecture named \modelname that addresses the common issue in spatio-temporal inference of choosing a specific order and assuming independence in spatial and temporal trends. We do this by decoupling inference to spatio-temporal and temporo-spatial inference and intelligently combining the outputs of both permutations. We extend \modelname to apply to the incident prediction problem in road traffic systems that follow daily and weekly periodicity. Using historic averages and predictions from \modelname, we generate incident labels and compare those against the ones generated using the ground-truth data. Our evaluations demonstrate a superior performance of \modelname for both incident detection and diagnosis for two publicly available highway traffic datasets and a new non-highway dataset. Future work will investigate extensions to dynamically adapt \modelname in case of non-stationary graph inputs, such as in cases of road or lane closures.

\section*{Acknowledgements}
We thank Transport for Greater Manchester (TfGM) for providing us with the raw data signals of the road-traffic network in the Radcliffe area. Work done by first two authors as interns at Transport Research Laboratory - Software, UK in partnership with the Alan Turing Institute. Shreshth Tuli is supported by the President's PhD Scholarship at Imperial College London. 

\bibliographystyle{named}
\bibliography{ijcai22}

\appendix

\section{Dataset Collection and Feature Details}
\label{sec:datset_details}

\begin{figure}
    \centering
    \begin{subfigure}{0.575\linewidth}
        \centering
        \includegraphics[width=\linewidth]{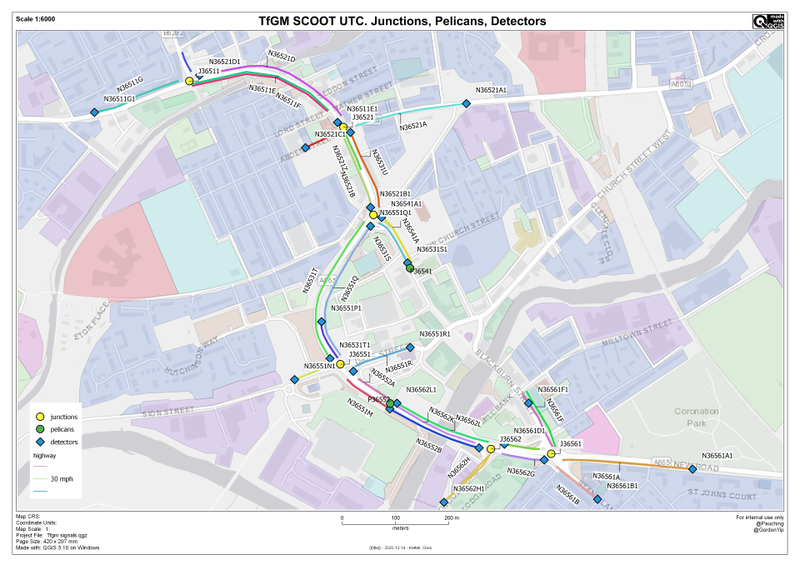}
        \caption{Radcliffe Sensor Distribution}
        \label{fig:radcliffe_map}
    \end{subfigure}
    \begin{subfigure}{0.405\linewidth}
        \centering
        \includegraphics[width=\linewidth]{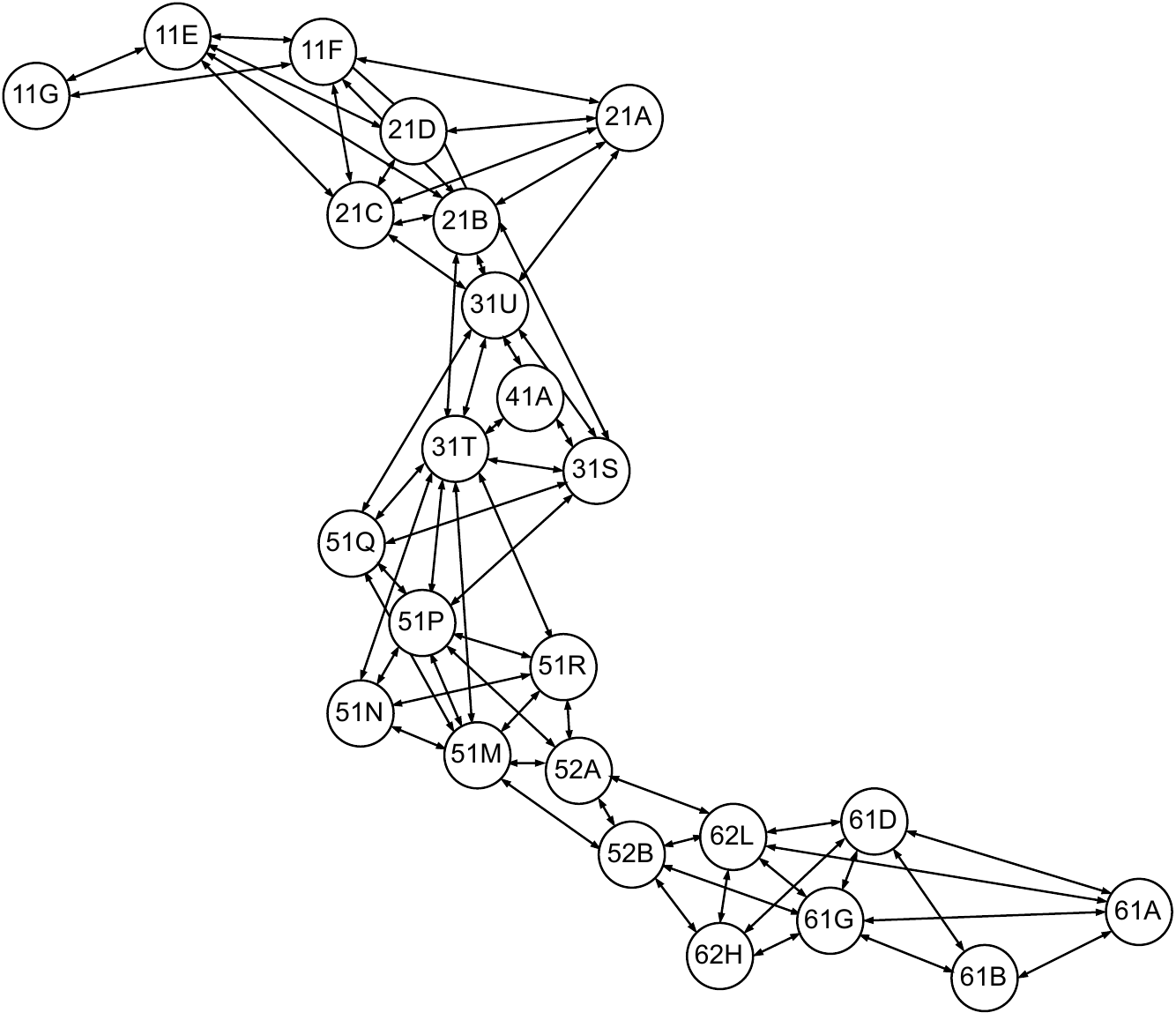}
        \caption{Graph}
        \label{fig:radcliffe_graph}
    \end{subfigure}
    \caption{Map of the Radcliffe Area for the collection of \dataset Dataset and the corresponding graph with \textit{junctions} as nodes and \textit{links} as edges. The \textit{junctionIDs} on the left are truncated on the right to remove the common token (eg., N36511E as 11E). Zoom to read.}
    \label{fig:radcliffe}
\end{figure}

The \dataset dataset is a spatio-temporal dataset collected from the Open Data Portal of TRL SCOOT UTC~\cite{scoot} traffic monitoring and control system. The data involved the Radcliffe area, see Figure~\ref{fig:radcliffe}. The data is obtained from induction loops installed on links that detect the presence of vehicles, pedestrians and other road users. This data is then passed to the SCOOT model that reports traffic metrics at quarter-second intervals. Using this data, the SCOOT system maintains a model of the road network. This model leverages the detector data to produce a vehicle arrival process at each link of the traffic network. The model then is used to generate features such as traffic flow rate and congestion index. The features collected for each link are described below:
\begin{enumerate}
    \item \texttt{del}: delay (seconds) experienced by the vehicles at the stopline of each link.
    \item \texttt{flow}: flow rate (vehicles per hour) generated by the SCOOT model.
    \item \texttt{flow\_raw}: raw flow (vehicles) detected in each traffic signal cycle.
    \item \texttt{cong}: congestion (inter-vehicle interval time) generated by the SCOOT model.
    \item \texttt{cong\_raw:} raw congestion (inter-vehicle interval time) detected by each loop sensor.
    \item \texttt{occ}: occupancy of the link generated by the SCOOT model.
    \item \texttt{ql}: queue length (vehicles) at link's stopline generated by the model.
\end{enumerate}

\textit{Possible sources of errors.} 
The SCOOT model is unaware of the traffic queues or flow rates outside the modeled environment, which could give rise to inaccuracies in the model generated features. However, considering the lack of non-highway datasets, \dataset enables research for European non-highway roads, significantly different from American highway networks. 

\noindent
\textbf{Dataset Availability.} The processed baselines and datasets are available at Zenodo: \url{https://zenodo.org/record/6524777}.

\section{Model Architecture}
\label{sec:arch}
We detail the implementation of the \modelname architecture introduced in the paper and shown in Figure~\ref{fig:nn}.  
\begin{itemize}
    \item \textit{Transformer:} We use a dropout value of 0.1 in the position encoder, the transformer encoder and decoder layers. We use a single hidden layer of size 16 for both transformer encoder and decoders. The number of heads was kept same as the number of features (dimension size) of the datasets, \textit{i.e.}, one for METR-LA and PEMS and seven for \dataset.
    \item \textit{Graph Attention Network (GAT):} We use two graph attention networks. The number of heads for these networks was kept to be one. Thus, the output of the GAT in spatio-temporal inference shown in equation~\eqref{eq:spatio_temporal} had the channel size that is same size as window size $K$, giving $K$ encodings of the window of feature matrices. For the temporo-spatial inference shown in equation~\eqref{eq:temporo_spatial}, we get an output with a single feature matrix. 
    \item \textit{Weighted Skip Connection:} To generate the attention weights for the three encodings: spatio-temporal, temporo-spatial and input feature matrix at timestep $t$, we use a feed-forward model. This was a single layer neural network with input of size $N \times D$ and output of size 3 with the $\mathrm{softmax}$ activation function. 
    \item \textit{Feed-Forward Decoder:} The output after the weighted-skip connection was sent to a feed-forward decoder with 2 hidden layers, each of size 64 and the $\mathrm{LeakyReLU}$ activation function.
\end{itemize}

\section{Hyperparameter Details}
\label{sec:hyperparam}

For the \modelname model, we use a learning rate of $5\times10^{-4}$ for \dataset and METR-LA and $2\times10^{-4}$. We also use a weight decay of $10^{-5}$ to avoid over-fitting. We use a window size of $K = 5$ for all datasets as per grid search. For POT parameters, we set the percentile values as 99, 50 and 45 for \dataset, METR-LA and PEMS respectively for the 5 minute time horizon. As we increase the time horizon, we decrease the percentile values by $\delta = 0.5$ for \dataset and 2.5 for METR-LA and PEMS respectively. This is due to the noisy predictions as the time horizon increases, giving rise to frequent over-estimations and increasing the POT value. All hyperparameters, including $K$, $\delta$ and POT percentiles, were determined using grid-search aiming to maximize the F1 score on the validation set.

\begin{table}[t]
    \centering
\begin{tabular}{@{}lccc@{}}
\toprule 
Model & METR-LA & PEMS & \dataset\tabularnewline
\midrule
DCRNN & 0.68 & 1.63 & 0.45\tabularnewline
GTS & 0.44 & 1.31 & 0.29\tabularnewline
GCN & 0.47 & 1.08 & 0.31\tabularnewline
HAGCN & 0.27 & 0.65 & 0.18\tabularnewline
LSTM & 0.20 & 0.54 & 0.13\tabularnewline
TSE\_SC & 0.66 & 1.44 & 0.44\tabularnewline
DKFN & 1.01 & 2.13 & 0.67\tabularnewline
GraphWaveNet & 4.68 & 11.93 & 3.12\tabularnewline
\modelname & 1.16 & 2.93 & 0.77\tabularnewline
\bottomrule
\end{tabular}
    \caption{Number of parameters (in millions) of \modelname and state-of-the-art methods for each dataset.}
    \label{tab:params}
\end{table}

\begin{figure}[t]
    \centering
    \includegraphics[width=\linewidth]{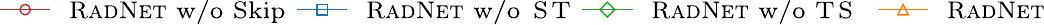}\\ 
    \vspace{3pt}
    \begin{subfigure}{0.325\linewidth}
        \centering
        \includegraphics[width=\linewidth]{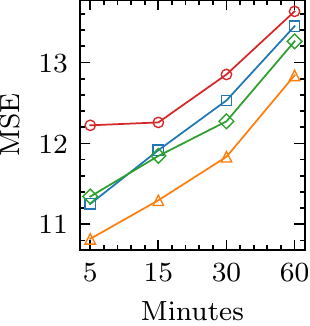}
        \caption{MSE}
        \label{fig:metrla_rmse}
    \end{subfigure}
    \begin{subfigure}{0.325\linewidth}
        \centering
        \includegraphics[width=\linewidth]{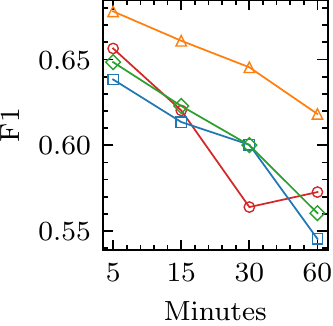}
        \caption{F1}
        \label{fig:metrla_f1}
    \end{subfigure}
    \begin{subfigure}{0.325\linewidth}
        \centering
        \includegraphics[width=\linewidth]{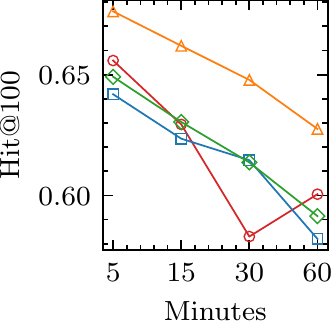}
        \caption{HitRate@100\%}
        \label{fig:metrla_h}
    \end{subfigure}
    \caption{Ablation Analysis on the METR-LA Dataset.}
    \label{fig:ablation_metrla}
\end{figure}

\begin{figure}[t]
    \centering
    \includegraphics[width=\linewidth]{images/ablations/legend.pdf}\\ 
    \vspace{3pt}
    \begin{subfigure}{0.325\linewidth}
        \centering
        \includegraphics[width=\linewidth]{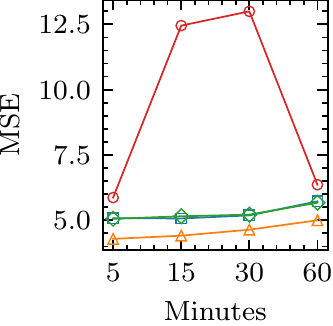}
        \caption{MSE}
        \label{fig:pems_rmse}
    \end{subfigure}
    \begin{subfigure}{0.325\linewidth}
        \centering
        \includegraphics[width=\linewidth]{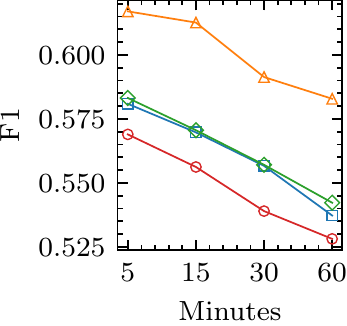}
        \caption{F1}
        \label{fig:pems_f1}
    \end{subfigure}
    \begin{subfigure}{0.325\linewidth}
        \centering
        \includegraphics[width=\linewidth]{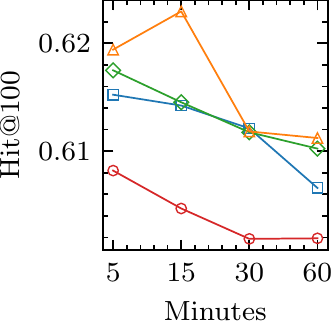}
        \caption{HitRate@100\%}
        \label{fig:pems_h}
    \end{subfigure}
    \caption{Ablation Analysis on the PEMS Dataset.}
    \label{fig:ablation_pems}
\end{figure}

\begin{figure}[!t]
    \centering
    \includegraphics[width=\linewidth]{images/ablations/legend.pdf}\\ 
    \vspace{3pt}
    \begin{subfigure}{0.325\linewidth}
        \centering
        \includegraphics[width=\linewidth]{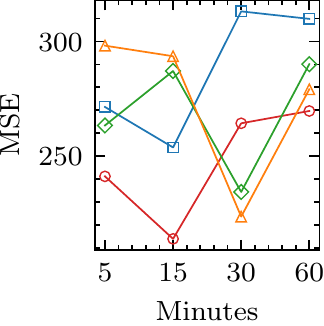}
        \caption{MSE}
        \label{fig:scoot_rmse}
    \end{subfigure}
    \begin{subfigure}{0.325\linewidth}
        \centering
        \includegraphics[width=\linewidth]{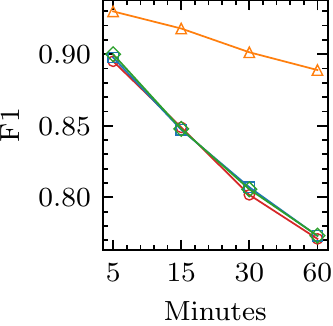}
        \caption{F1}
        \label{fig:scoot_f1}
    \end{subfigure}
    \begin{subfigure}{0.325\linewidth}
        \centering
        \includegraphics[width=\linewidth]{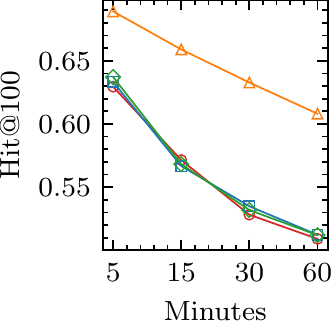}
        \caption{HitRate@100\%}
        \label{fig:scoot_h}
    \end{subfigure}
    \caption{Ablation Analysis on the \dataset Dataset.}
    \label{fig:ablation_scoot}
\end{figure}

\section{Additional Results}
\label{sec:additional_results}

\begin{table}[t]
    \centering
\resizebox{\linewidth}{!}{
\begin{tabular}{@{}lcccccccc@{}}
\toprule 
Horizon & Model & P & R & F1 & H@1 & H@1.5 & N@1 & N@1.5\tabularnewline
\midrule 
\multicolumn{9}{c}{\dataset}\tabularnewline
\midrule 
\multirow{2}{*}{5 min} & \modelname{*} & 0.931 & 0.938 & 0.935 & 0.679 & 0.722 & 0.679 & 0.719\tabularnewline
 & \modelname & 0.925 & 0.935 & 0.930 & 0.689 & 0.750 & 0.685 & 0.740\tabularnewline
\midrule
\multirow{2}{*}{15 min} & \modelname{*} & 0.873 & 0.948 & 0.913 & 0.658 & 0.738 & 0.654 & 0.726\tabularnewline
 & \modelname & 0.889 & 0.949 & 0.918 & 0.659 & 0.762 & 0.653 & 0.741\tabularnewline
\midrule
\multirow{2}{*}{30 min} & \modelname{*} & 0.851 & 0.948 & 0.896 & 0.622 & 0.718 & 0.620 & 0.700\tabularnewline
 & \modelname & 0.857 & 0.950 & 0.901 & 0.633 & 0.742 & 0.626 & 0.722\tabularnewline
\midrule
\multirow{2}{*}{60 min} & \modelname{*} & 0.830 & 0.922 & 0.874 & 0.585 & 0.685 & 0.577 & 0.662\tabularnewline
 & \modelname & 0.827 & 0.961 & 0.889 & 0.608 & 0.716 & 0.599 & 0.694\tabularnewline
\midrule
\multicolumn{9}{c}{METR-LA}\tabularnewline
\midrule
\multirow{2}{*}{5 min} & \modelname{*} & 0.615 & 0.616 & 0.616 & 0.614 & 0.806 & 0.614 & 0.780\tabularnewline
 & \modelname & 0.679 & 0.677 & 0.678 & 0.676 & 0.840 & 0.676 & 0.816\tabularnewline
\midrule
\multirow{2}{*}{15 min} & \modelname{*} & 0.676 & 0.641 & 0.658 & 0.658 & 0.826 & 0.661 & 0.807\tabularnewline
 & \modelname & 0.679 & 0.644 & 0.661 & 0.662 & 0.833 & 0.664 & 0.809\tabularnewline
\midrule
\multirow{2}{*}{30 min} & \modelname{*} & 0.690 & 0.620 & 0.653 & 0.656 & 0.827 & 0.660 & 0.807\tabularnewline
 & \modelname & 0.681 & 0.613 & 0.645 & 0.648 & 0.820 & 0.652 & 0.803\tabularnewline
\midrule
\multirow{2}{*}{60 min} & \modelname{*} & 0.702 & 0.599 & 0.646 & 0.650 & 0.818 & 0.656 & 0.806\tabularnewline
 & \modelname & 0.671 & 0.573 & 0.618 & 0.628 & 0.808 & 0.633 & 0.792\tabularnewline
\midrule 
\multicolumn{9}{c}{PEMS}\tabularnewline
\midrule
\multirow{2}{*}{5 min} & \modelname{*} & 0.604 & 0.599 & 0.601 & 0.604 & 0.847 & 0.604 & 0.814\tabularnewline
 & \modelname & 0.620 & 0.614 & 0.617 & 0.619 & 0.853 & 0.619 & 0.821\tabularnewline
\midrule
\multirow{2}{*}{15 min} & \modelname{*} & 0.616 & 0.584 & 0.599 & 0.610 & 0.848 & 0.611 & 0.817\tabularnewline
 & \modelname & 0.630 & 0.596 & 0.613 & 0.623 & 0.855 & 0.624 & 0.824\tabularnewline
\midrule
\multirow{2}{*}{30 min} & \modelname{*} & 0.616 & 0.556 & 0.584 & 0.606 & 0.851 & 0.606 & 0.815\tabularnewline
 & \modelname & 0.624 & 0.562 & 0.591 & 0.612 & 0.851 & 0.613 & 0.818\tabularnewline
\midrule
\multirow{2}{*}{60 min} & \modelname{*} & 0.597 & 0.515 & 0.553 & 0.581 & 0.829 & 0.587 & 0.806\tabularnewline
 & \modelname & 0.631 & 0.542 & 0.583 & 0.611 & 0.850 & 0.613 & 0.819\tabularnewline
\bottomrule
\end{tabular}
}
    \caption{Comparison between the direct (\modelname) and autoregressive (\modelname{*}) forecasting for incident prediction.}
    \label{tab:autoregressive}
\end{table}

\noindent \textbf{Model Size Comparison.}
Table~\ref{tab:params} compares the size of the parameter sets of the \modelname neural network with those of the benchmark methods. Compared to the best baseline, \textit{i.e.}, GraphWaveNet, \modelname has lower model size. This shows that the \modelname model gives higher performance without a high computational footprint.

\noindent \textbf{Ablation Analysis.} To study the relative importance of each component of the model, we exclude every major one and observe how it affects the performance of the \modelname model for each dataset. Figures~\ref{fig:ablation_scoot},~\ref{fig:ablation_metrla} and~\ref{fig:ablation_pems} compare the converged mean square error (loss function), F1 score and HitRate@100\% for \modelname and the following variants:
\begin{itemize}
    \item \modelname w/o Skip: The \modelname without the weighted-skip connection. Instead, the two outputs and the input feature matrix are added and forwarded to the feed-forward decoder.
    \item \modelname w/o ST: The model without the spatio-temporal inference; thus, having a single instance of the GAT and Transformer networks. The weighted skip is between $W_1^{(t)}$ and $X^{(t)}$, having the $w$ vector of size 2.
    \item \modelname w/o TS: The model without the temporo-spatial inference, again with a single instance of the GAT and Transformer networks. The weighted skip is between $W_2^{(t)}$ and $X^{(t)}$.
\end{itemize}
Hyperparameter exploration was performed independently for the ablation models. The results demonstrate that all components are critical to achieving the best performance. Specifically, when the weighted-skip connection is replaced with a simple addition of the encoded outputs and the input matrix, the performance drop is the highest. This corroborates our claims of the requirement of dynamic importance to the disparate inference permutations to account for the non-stationary behaviors of road traffic networks.

\noindent \textbf{Model Exploration to Autoregressive Prediction.} Akin to the benchmark methods, the \modelname is trained and tuned separately for each time horizon. However, this makes the training approach specific to the horizon value and limits us from forecasting for a new time horizon without training the model from scratch. To address this, we develop a single-step forecasting method that aims to generate the feature matrix and call this \modelname{*}. Thus, to generate a forecast for any given horizon value $H$, we train a model $f_\theta(.)$ that forecasts
\begin{equation*}
    \hat{X}^{(t+1)} = f_\theta(W^{(t)}, G).
\end{equation*}
Now, we use the new window for the next step and generate a two-step forecast autoregressively as 
\begin{align*}
    \hat{W}^{(t+1)} &= X^{(t-K+1):(t)}, \hat{X}^{(t+1)},\\
    \hat{X}^{(t+2)} &= f_\theta(\hat{W}^{(t+1)}, G).
\end{align*}
We can continue this for $H$ steps to generate $X^{(t+H)}$. As this is agnostic to the time horizon and autoregressively generates single-step forecasts, this model needs to be trained only once to perform predictions for different horizon values. The results for the \modelname* against \modelname are shown in Table~\ref{tab:autoregressive}. The \modelname{*} is trained using autoregressive predictions till 15 minutes horizon at training time and teacher forcing where we set $W^{(t+i)}$ (where $i \in \{1, \ldots, H\}$) from the dataset instead of utilizing $\hat{W}^{(t+i)}$ forecasted by the model. In training, we use a constant teacher-forcing probability of $p = 0.2$ of using the forecasted feature window to iteratively update the input instead of using ground-truth window. Although the \modelname{*} gives lower scores in general compared to \modelname, the performance is higher for 30 and 60 minute horizon values in the case of the METR-LA dataset. Nevertheless, the key advantage of reduced training time calls for further investigation as part of future work.
\end{document}